\documentclass[english]{article}
\usepackage[T1]{fontenc}
\usepackage[utf8]{inputenc}
\usepackage{float}
\usepackage{booktabs}
\usepackage{amsmath}
\usepackage{amssymb}
\usepackage{graphicx}
\usepackage{microtype}

\makeatletter

\providecommand{\tabularnewline}{\\}

\usepackage[nonatbib, final]{nips_2018}

\usepackage[utf8]{inputenc} 
\usepackage[T1]{fontenc}    
\usepackage{hyperref}       
\usepackage{url}            
\usepackage{booktabs}       
\usepackage{amsfonts}       
\usepackage{nicefrac}       

\makeatother

\usepackage{babel}
\begin{document}
\title{Text Generation with Deep Variational GAN}
\author{Mahmoud Hossam$^{1}$, Trung Le$^{1}$, Michael Papasimeon$^{2}$,
Viet Huynh$^{1}$, Dinh Phung$^{1}$ \\ $^{1}$Faculty of Information
Technology, Monash University\\ Clayton, VIC 3800 \\ $^{2}$School
of Computing and Information Systems, The University of Melbourne\\
Parkville, VIC 3052 \\ \texttt{$^{1}$\{mhossam, trunglm, viet.huynh,
dinh.phung\}@monash.edu}\\\texttt{$^{2}$michael.papasimeon@unimelb.edu.au}}

\maketitle
\global\long\def\mF{\mathcal{F}}%

\global\long\def\mA{\mathcal{A}}%

\global\long\def\mH{\mathcal{H}}%

\global\long\def\mX{\mathcal{X}}%

\global\long\def\dist{d}%

\global\long\def\HX{\entro\left(X\right)}%
 
\global\long\def\entropyX{\HX}%

\global\long\def\HY{\entro\left(Y\right)}%
 
\global\long\def\entropyY{\HY}%

\global\long\def\HXY{\entro\left(X,Y\right)}%
 
\global\long\def\entropyXY{\HXY}%

\global\long\def\mutualXY{\mutual\left(X;Y\right)}%
 
\global\long\def\mutinfoXY{W\mutualXY}%

\global\long\def\given{\mid}%

\global\long\def\gv{\given}%

\global\long\def\goto{\rightarrow}%

\global\long\def\asgoto{\stackrel{a.s.}{\longrightarrow}}%

\global\long\def\pgoto{\stackrel{p}{\longrightarrow}}%

\global\long\def\dgoto{\stackrel{d}{\longrightarrow}}%

\global\long\def\ll{\mathit{l}}%

\global\long\def\logll{\mathcal{L}}%

\global\long\def\bzero{\vt0}%

\global\long\def\bone{\mathbf{1}}%

\global\long\def\bff{\vt f}%

\global\long\def\bx{\boldsymbol{x}}%

\global\long\def\bX{\boldsymbol{X}}%

\global\long\def\bW{\mathbf{W}}%

\global\long\def\bH{\mathbf{H}}%

\global\long\def\bL{\mathbf{L}}%

\global\long\def\tbx{\tilde{\bx}}%

\global\long\def\by{\boldsymbol{y}}%

\global\long\def\bY{\boldsymbol{Y}}%

\global\long\def\bz{\boldsymbol{z}}%

\global\long\def\bZ{\boldsymbol{Z}}%

\global\long\def\bu{\boldsymbol{u}}%

\global\long\def\bU{\boldsymbol{U}}%

\global\long\def\bv{\boldsymbol{v}}%

\global\long\def\bV{\boldsymbol{V}}%

\global\long\def\bw{\vt w}%

\global\long\def\balpha{\gvt\alpha}%

\global\long\def\bbeta{\gvt\beta}%

\global\long\def\bmu{\gvt\mu}%

\global\long\def\btheta{\boldsymbol{\theta}}%

\global\long\def\blambda{\boldsymbol{\lambda}}%

\global\long\def\realset{\mathbb{R}}%

\global\long\def\realn{\real^{n}}%

\global\long\def\natset{\integerset}%

\global\long\def\interger{\integerset}%

\global\long\def\integerset{\mathbb{Z}}%

\global\long\def\natn{\natset^{n}}%

\global\long\def\rational{\mathbb{Q}}%

\global\long\def\realPlusn{\mathbb{R_{+}^{n}}}%

\global\long\def\comp{\complexset}%
 
\global\long\def\complexset{\mathbb{C}}%

\global\long\def\and{\cap}%

\global\long\def\compn{\comp^{n}}%

\global\long\def\comb#1#2{\left({#1\atop #2}\right) }%

\begin{abstract}
Generating realistic sequences is a central task in many machine learning
applications. There has been considerable recent progress on building
deep generative models for sequence generation tasks. However, the
issue of mode-collapsing remains a main issue for the current models.
In this paper we propose a GAN-based generic framework to address
the problem of mode-collapse in a principled approach. We change the
standard GAN objective to maximize a variational lower-bound of the
log-likelihood while minimizing the Jensen-Shanon divergence between
data and model distributions. We experiment our model with text generation
task and show that it can generate realistic text with high diversity.
\end{abstract}

\section{Introduction\label{sec:intro}}

Realistic sequence generation is one of the most important tasks in
machine learning. In many applications such as natural language processing,
music synthesis, biological sequences design, robotics, and dynamical
systems modeling, a model that is able to learn in an unsupervised
manner from data is crucial. Recently, there has been a considerable
amount of work on developing deep generative models to generate discrete
sequences of text using adversarial learning \cite{yu2017seqgan,zhang2017adversarial,chen2018adversarial,fedus2018maskgan}.
However, two fundamental problems emerge in these models; non-differentiability
of discrete data and mode-collapsing -- the lack of ability to capture
various modes of data. 

Current approaches to learning sequence generation usually focus on
tackling the non-differentiabiliy obstruction, and can be grouped
into two main approaches. The first group of models makes use of reinforcement
learning techniques like policy gradients, to overcome the non-differentiability
with discrete data. In these architectures, the discriminator networks
(fake/real binary classifiers) are disconnected from the generator
networks (the main networks), and policy gradients are used to estimate
an error signal from the discriminator \cite{yu2017seqgan,fedus2018maskgan}.
The main issue with these techniques is the high variance of the gradients
estimations \cite{chen2018adversarial}. In addition, they do not
incorporate latent space learning, which allows learning higher representations
of the data. The other group of models still employ a fully differentiable
GAN \cite{goodfellow2014generative} network, but they make use of
Gumbel Softmax trick \cite{zhang2017adversarial,chen2018adversarial}
to overcome the non-differentiability problem.

However, the issue of mode-collapsing has not been addressed yet in
a principled way in all of the aforementioned models. In this paper,
we present a principled way to alleviate the mode collapsing problem
of sequence generation models and apply our framework to text generation
task. While our work is demonstrated with discrete data, it can be
straightforwardly adopted for continuous data.

In practice, GANs usually suffer from ``mode-collapsing'' problem
\cite{metz2016unrolled,poole2016improved,hoang2018mgan}, where the
generator learns to map several different $z$ values to the same
output point, relying on a few modes from the data distribution. This
causes GAN to be incapable of generating diverse samples from the
given latent codes. In this paper, we propose a new framework called
Adversarial Auto-regressive Networks (ARN) to address the mode collapsing
problem in a principled way when generating sequences using adversarial
training. The main highlights of our model include: (i) the capability
of learning to generate sentences either from random latent space
or conditioned on the first token; (ii) using standard back-propagation
with relaxation instead of policy gradients; (iii) overcoming mode-collapsing
issues and achieving high diversity scores.

\section{Adversarial Autoregressive Networks (ARN) \label{sec:model}}

\global\long\def\E{\mathbb{E}}%

\global\long\def\var{\mathbb{V}}%

\begin{figure}[H]
\begin{centering}
\includegraphics[width=0.75\textwidth]{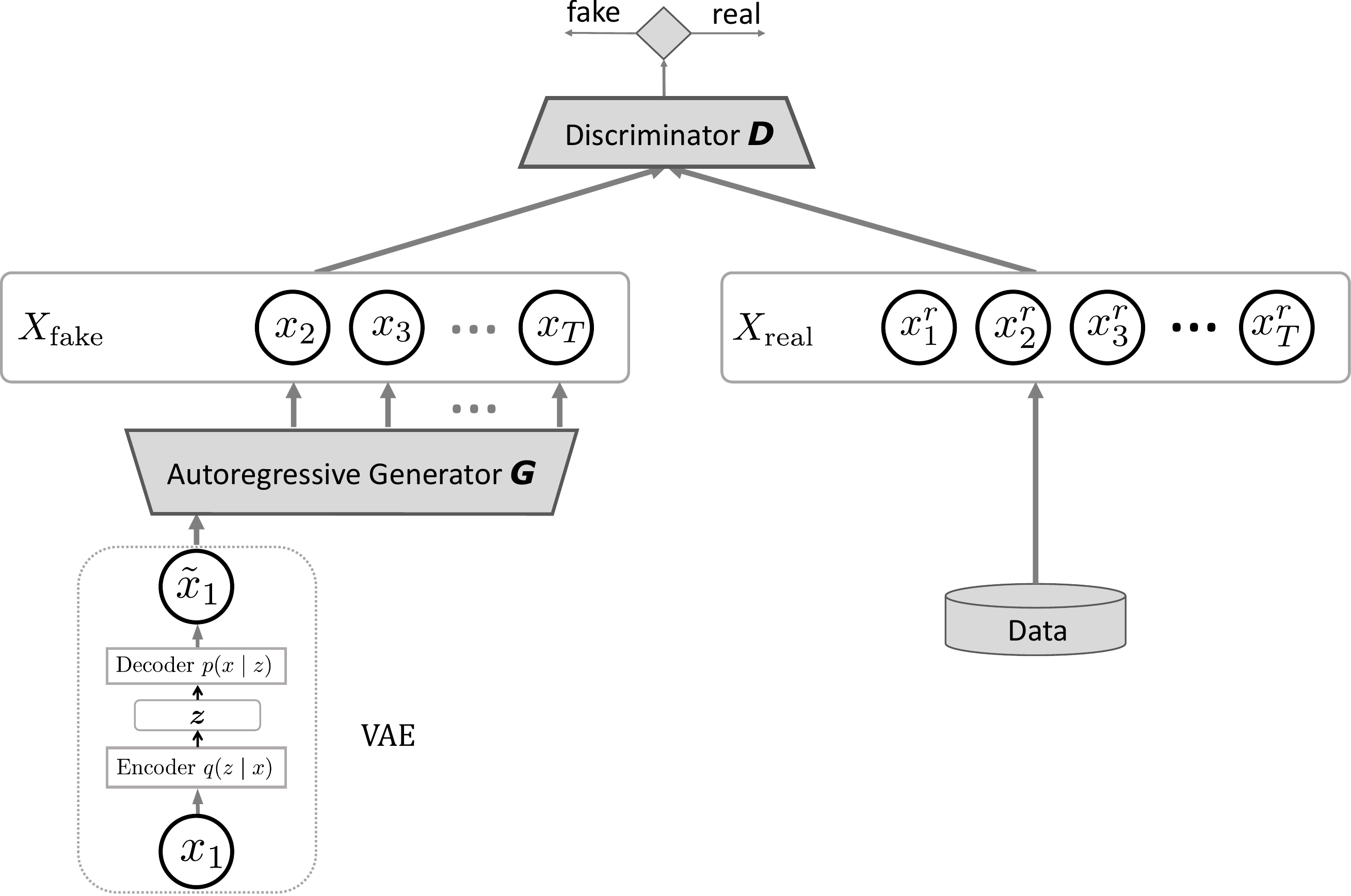}\caption{\label{fig:ARN}Proposed Adversarial Autoregressive Network}
\par\end{centering}
\end{figure}
Adversarial Autoregressive Network is mainly built using an autoregressive
generator, like RNN or LSTM, trained in a GAN framework for sequence
generation. However, in order to learn a latent space that can be
used to control the sequence generation, we employ a variational autoencoder
at the first token, $x_{1}$ (Figure \ref{fig:ARN}). Below we discuss
in details the derivation and intuition behind our model.

\textbf{Model definition.} A sample $X$ in our setting is defined
as a sequence of $T$ tokens denoted by $X=\left[x_{1},x_{2},...,x_{T}\right]$,
where we assume that all samples have length $T$. For our autoregressive
model with model parameters $\theta$, the log-likelihood can be written
as:
\begin{align*}
\log\,p_{G}(X\mid\theta) & =\sum_{i=2}^{T}\log\,p_{G}(x_{i}|\,h_{i-1},\theta)\,+\,\log\,p_{G}(x_{1}\mid\theta),
\end{align*}
\vspace{-3mm}

This is the default neural autoregressive model formulation. Now we
start introducing an adversarial learning framework for this model
by introducing a latent variable $z$ to the autoregressive model,
where we rewrite $\log\,p(x_{1}\mid\theta)$ as marginalization over
the $z$:{\small{}
\begin{align}
\log\,p_{G}(x_{1}\mid\theta) & =\log\,\sum_{z}\,p_{G}(x_{1},\,z\mid\theta)\geqslant-I_{KL}(q(z\mid x_{1},\phi)\,||\,p(z))+\E_{q(z\mid x_{1},\phi)}[\log\,p_{G}(x_{1}|\,z,\theta)],\label{eq:1}
\end{align}
}where $I_{KL}$ is Kullback--Leibler divergence, $q(z\mid x_{1},\phi)$
is an approximation of the posterior $p\left(z\mid x_{1},\theta\right)$
and $p(z)$ is a prior distribution to $z$. The right hand side of
Eq. (\ref{eq:1}) is a lower bound for $\log\,p_{G}(x_{1}\mid\theta)$.
We can then write $\log\,p_{G}(X\mid\theta)$ in terms of a lower
bound as:{\small{}
\begin{align}
\log\,p_{G}(X & \mid\theta)\geqslant\sum_{i=2}^{T}\log\,p_{G}(x_{i}|\,h_{i-1},\theta)-I_{KL}(q(z\mid x_{1},\phi)\,\Vert\,p(z))+\E_{q(z\mid x_{1},\phi)}[\log\,p_{G}(x_{1}|\,z,\theta)].\label{eq:2}
\end{align}
}{\small\par}

We propose to incorporate adversarial learning to autoregressive
sequential model in a principled way. One generator $G\left(z\right)$
and one discriminator $D\left(X\right)$ are employed to create a
game like in GAN while the task of the discriminator is to discriminate
true data and fake data and the task of the generator is to generate
fake data that maximally make the discriminator confused. In addition,
the generator $G$ is already available which departs from a noise
$z\sim p_{z}$, uses the conditional distribution $p\left(x_{1}\mid z,\theta\right)$
to generate $x_{1}$, and follows the autoregressive model to consecutively
generate $x_{2:T}$. We come with the following minimax problem:
\begin{align}
\max_{G}\min_{D}\,[ & \mathbb{E}_{X\sim p_{d}}\left[\log\,p_{G}\left(X\mid\theta\right)\right]-\text{\ensuremath{\mathbb{E}}}_{X\sim p_{d}}\left[\log\,D\left(X\right)\right]-\mathbb{E}_{z\sim p_{z}}\left[\log\left[1-D\left(G\left(z\right)\right)\right]\right]\,],\label{eq:op_ARN}
\end{align}
where the generator $G$ consists of the decoder $p\left(x_{1}\mid z,\theta\right)$,
the autoregressive model, hence $G$ is parameterized by $\left(\theta,\phi\right)$,
and $\log\,p_{G}\left(X\mid\theta\right)$ is substituted by its lower
bound in Eq. (\ref{eq:2}).We can theoretically prove that the minimax
problem in Eq. (\ref{eq:op_ARN}) is equivalent to the following optimization
problem (see the proof in Appendix):
\begin{equation}
\min_{G}\,I_{KL}\left(P_{d}\,||\,P_{G}\right)+I_{JS}\left(P_{d}\,||\,P_{G}\right),\label{eq:op_diver}
\end{equation}
where $I_{JS}$ is Jenshen-Shannon divergence and $P_{G}$ is the
generative distribution. The optimization problem in Eq. (\ref{eq:op_diver})
reveals that at the Nash equilibrium point the generative distribution
$P_{G}$ is exactly the data distribution $P_{d}$, thus overcoming
the mode-collapse issue caused by original GAN formulation.

To train our model, we alternatively update $G$ and $D$ with relevant
terms. We note that in the optimization for updating $G$ regarding
$\log\,p_{G}\left(X\mid\theta\right)$, we maximize its lower bound
in Eq. (\ref{eq:2}) instead of the likelihood function. In addition,
to overcome the non-differentiability of discrete data in our model,
we use Gumbel Softmax relaxation \cite{DBLP:journals/corr/JangGP16,DBLP:journals/corr/MaddisonMT16}.

\section{Experiments \label{sec:Experiments}}

\begin{table}
\caption{\label{tab:IMDB}BLEU, FC and Diversity scores}

\centering{}{\small{}}%
\begin{tabular}{ccccccc}
\cmidrule{2-7} \cmidrule{3-7} \cmidrule{4-7} \cmidrule{5-7} \cmidrule{6-7} \cmidrule{7-7} 
 & \multicolumn{6}{c}{{\small{}IMDB Reviews}}\tabularnewline
\cmidrule{2-7} \cmidrule{3-7} \cmidrule{4-7} \cmidrule{5-7} \cmidrule{6-7} \cmidrule{7-7} 
 & {\small{}BLEU-2 \%} & {\small{}BLEU-3 \%} & {\small{}FC-2 \%} & {\small{}FC-3 \%} & {\small{}Diversity-2 \%} & {\small{}Diversity-3 \%}\tabularnewline
\midrule
{\small{}SeqGAN} & \textbf{\small{}86.38} & \textbf{\small{}54.19} & {\small{}11.00} & {\small{}11.98} & {\small{}21.14} & {\small{}49.98}\tabularnewline
{\small{}Ours (decoded $x_{1}$)} & {\small{}69.04} & {\small{}30.70} & \textbf{\small{}14.28} & \textbf{\small{}12.04} & \textbf{\small{}31.51} & \textbf{\small{}64.21}\tabularnewline
{\small{}Ours (noise)} & {\small{}69.23} & {\small{}30.66} & \textbf{\small{}14.23} & \textbf{\small{}12.11} & \textbf{\small{}31.10} & \textbf{\small{}64.13}\tabularnewline
\end{tabular}{\small\par}
\end{table}
We train our model for text generation task using the IMDB movie reviews
dataset \cite{DBLP:conf/acl/MaasDPHNP11}. The dataset consists of
100,000 reviews extracted from IMDB website. The training set is divided
into three groups; positive, negative (12,500 reviews each) and unlabeled
reviews (50,000), and the testing set is 12,500 positive and negative
each. In all of our experiments, we use sentence length of 20 words
and vocabulary size of 10000. We use 500 dimensional embedding vectors,
and 500 units hidden layer for LSTM Generator and 350 units for VAE.

We evaluate both the quality and mode-collapse of the generated sentences.
To evaluate quality, we use BLEU score \cite{DBLP:conf/acl/PapineniRWZ02},
which is commonly used in machine translation to compare the quality
of candidate translations compared to the ground-truth reference.
To evaluate the mode-collapse, we use two n-gram based scores inspired
by \cite{fedus2018maskgan}, namely the \textit{Diversity }and ``\textit{Feature
Coverage} (\textit{FC})'' scores.

\textbf{Diversity Score} We define the diversity as the ability to
generate sentences with diverse n-grams that are not necessarily found
in the test set. We measure this by computing the percentage of unique
n-grams generated by the model relative to the number of all n-grams
generated by the model. 

\textbf{Feature Coverage (}\textbf{\textit{FC}}\textbf{) Score} The
\textit{FC} score is used to measure how well the model covers all
features (n-grams) of the data. The score is computed as the percentage
of unique n-grams generated by the model that is found in the test
set, relative to the number of all n-grams generated by the model. 

It is important to notice that \textit{FC} score doesn't necessarily
correlate with BLEU score, as it computes only the percentages of
``unique'' n-grams that match with the test set. This means that
unlike BLEU score, \textit{FC} score is affected by the diversity
of the generated sentences, where the higher the unique n-grams count,
the higher the \textit{FC} score, and vice versa. Sentences can be
generated from our model in two ways; by starting from a real first
word through the decoder, or from noise input through $z$. We compare
our model to SeqGAN \cite{yu2017seqgan} and report the results in
Table \ref{tab:IMDB} for 2-grams and 3-grams of all scores. 
\begin{table}
\caption{\label{tab:Generated-Sample-Sentences}Generated sample sentences}
\medskip{}

\centering{}{\footnotesize{}}%
\begin{tabular}{l}
\textbf{\footnotesize{}Our Model (starting from decoded $x_{1}$)}\tabularnewline
\midrule
{\footnotesize{}i had more interest for this movie . even though they
rightly watched it once . it's neither entertaining nor}\tabularnewline
\addlinespace[1pt]
{\footnotesize{}he just watched this show , one of the funniest moments
of <UNK> or plot . i watched this and}\tabularnewline
\addlinespace[1pt]
{\footnotesize{}victor is a zombie documentary of its own hype that
<UNK> puzzles on many of my favorites . and every}\tabularnewline
\addlinespace[1pt]
{\footnotesize{}i couldn't relate to this movie for the first time
only because time i watched the director whose friend i}\tabularnewline
\addlinespace[1pt]
\addlinespace
\textbf{\footnotesize{}Our Model (starting from noise $z$)}\tabularnewline
\midrule
{\footnotesize{}first movie was a letdown of a film . the first 30
minutes of it was ok , until i}\tabularnewline
\addlinespace[1pt]
{\footnotesize{}this movie is not only horrible as a bad movie , not
because it has been based on history .}\tabularnewline
\addlinespace[1pt]
{\footnotesize{}probably don't know i am a fan of horror movies ,
it was a little while ago and i think}\tabularnewline
\addlinespace[1pt]
{\footnotesize{}another is one of the worst movies , ranking up there
. the script is <UNK> by the film's predecessor}\tabularnewline
\addlinespace[1pt]
{\footnotesize{}the first review that i can't figure out what's a
<UNK> about this film when it came out in 1972}\tabularnewline
\addlinespace[1pt]
{\footnotesize{}this most awaited movie and it seems to be the worst
film ever . good plot , the storyline was}\tabularnewline
\addlinespace[1pt]
{\footnotesize{}holy movie is just fun . it's not to do justice .
the first monkey meets this hoping to be}\tabularnewline
\addlinespace[1pt]
{\footnotesize{}this film grabbed the attention to the plot with standard
<UNK> , and this never even saw a story with}\tabularnewline
\addlinespace[1pt]
{\footnotesize{}what i saw this episode too , i thought it was awesome
for several great actors . <br / science}\tabularnewline
\addlinespace[1pt]
{\footnotesize{}although movie <UNK> if not to forget with a british
tradition . i understand why the ideas presented well enough}\tabularnewline
\addlinespace[1pt]
{\footnotesize{}this 1985 cult film \textquotedbl{} animal \textquotedbl{}
was a <UNK> of a couple tells the most of youth , but}\tabularnewline
\addlinespace[1pt]
{\footnotesize{}this disappointing . . . richard murray was a landmark
in many two so far , such a pretentious crap}\tabularnewline
\addlinespace[1pt]
{\footnotesize{}people misguided . i found way almost no battlestar
<UNK> to complain about this movie ( and like a <UNK>}\tabularnewline
\addlinespace[1pt]
{\footnotesize{}this superbly finished the <UNK> , i was very excited
about martin carter and scott he ran the tv series}\tabularnewline
\addlinespace[1pt]
{\footnotesize{}this movie is really very sweet here . it reminded
me of the dvd both brought us that devil's <UNK> }\tabularnewline
\addlinespace[1pt]
{\footnotesize{}if just watched this movie when i was around 15 years
ago and although it looks like it was boring}\tabularnewline
\addlinespace[1pt]
\end{tabular}{\footnotesize\par}
\end{table}

We can see that ``Feature-Coverage'' score is very similar to the
baseline, while the Diversity score is higher. This suggests that
our model is capable of learning the same overall features (n-grams)
of the data, yet it is also able to generate more diverse samples
out of these learned features. To qualitatively evaluate the ``quality''
of this diverse output, we show generated samples with average or
low BLEU scores in Table \ref{tab:diverse_quality_samples}. We can
see that low BLEU sentences generally still have grammatical structure
and sometimes semantically meaningful, suggesting that the model does
not learn random or gibberish modes when they do not match directly
with data. In addition, we show general samples from the output in
Table \ref{tab:Generated-Sample-Sentences}.

While the BLEU scores are lower than the baseline, yet the \textit{FC}
and Diversity scores are higher. For text generation tasks, BLEU is
computed relative to the whole test corpus. This means that BLEU score
can increase significantly on the expense of output diversity, when
few generated sentences match highly with test corpus but are repeated
very frequently. Therefore, we see that it is essential that text
generative models are evaluated for both quality and diversity in
a unified manner \cite{fedus2018maskgan}.

\section{Conclusion\label{sec:conclusion}}

In this paper we presented a sequential deep generative model to generate
sequences based on a principled approach to address the mode collapse
problem. We applied the model to text generation task and showed that
the model can generate grammatically and semantically meaningful sentences
with high diversity.

In future work, we will investigate the latent space learning of the
model, aiming to learn smooth transitions between sentence styles
or sentiments. We intend to improve the latent learning using recent
methods such as learned similarity metric \cite{DBLP:journals/corr/LarsenSW15}
that include all of the input tokens, and incorporate the ability
to control output sentences conditioned on true data through the encoder.
We will also fine tune the model to achieve better results, design
more comprehensive quality/diversity measures, and compare with recent
sequential generative models \cite{zhang2017adversarial,chen2018adversarial}.

{\footnotesize{}\bibliographystyle{plain}
\bibliography{refs01}
}{\footnotesize\par}

\section*{Appendix}

\subsection*{Generative Adversarial Networks (GAN)}

\global\long\def\E{\mathbb{E}}%

\global\long\def\var{\mathbb{V}}%

The basic idea of Generative adversarial networks (GAN) \cite{goodfellow2014generative}
is the adversarial training between two players. The goal of the first
player, the generator $G$ , is to get very good at generating data
that is very close to the real data that comes from real distribution
$p_{d}(x)$. The goal of the second player, the discriminator $D$,
is to distinguish real data from fake data generated by the generator.
The standard GAN objective to optimize is the minimax game between
$D$ and $G$: 
\begin{equation}
\min_{G}\max_{D}\,\,\E_{x\sim p_{d}}\log\,D(x)\,+\,\E_{z\sim p_{z}}\log\,(1-D(G(z)))\label{eq:stdGAN}
\end{equation}
where $z$ is the random noise input to $G$, and $p_{z}$ is the
prior distribution of the $z$. After the training is finished, the
generator is used to generate data from any random input $z$.

\begin{table}
\caption{\label{tab:diverse_quality_samples}Low/average BLEU samples}

\centering{}%
\begin{tabular}{cl}
\toprule 
\textbf{\scriptsize{}BLEU-3} & \textbf{\scriptsize{}Sample sentences}\tabularnewline
\midrule
{\scriptsize{}00.00\%} & {\scriptsize{}a creative , well-made comedy that focuses into a scary
film by peter jackson , are heading for <UNK> }\tabularnewline
{\scriptsize{}00.00\%} & {\scriptsize{}this direction struck me , that's ! this dreadful zombie
movie , generally and curtis are terrific but doesn't . }\tabularnewline
{\scriptsize{}00.00\%} & {\scriptsize{}but effort , rural accent must be developed from joke
a video <UNK> this new version starts ! some say }\tabularnewline
{\scriptsize{}11.11\%} & {\scriptsize{}okay exceptionally documentary about steve hayes ( who
doesn't even destroy to the <UNK> ! that doesn't act better than }\tabularnewline
{\scriptsize{}11.11\%} & {\scriptsize{}sorry you're a fan , might even like your a spoiler
as a master of shakespeare ; a little girl }\tabularnewline
{\scriptsize{}11.11\%} & {\scriptsize{}the wish i can relate to seeing many kind of kills the
other well-made film adaptations than the godfather documentary }\tabularnewline
{\scriptsize{}22.22\%} & {\scriptsize{}wow hitchcock's prince meets thriller of the modern
relationships in <UNK> terror and <UNK> above <UNK> the war film and }\tabularnewline
{\scriptsize{}27.77\%} & {\scriptsize{}the fine ending is , not even close to \textquotedbl{}
princess <UNK> , \textquotedbl{} that it'd be extremely cool . }\tabularnewline
{\scriptsize{}27.77\%} & {\scriptsize{}this is an excellent , unbelievable , rather colorful
show . brian <UNK> did never become a straight to part }\tabularnewline
{\scriptsize{}33.33\%} & {\scriptsize{}brilliant carpenter's halloween is actually my favorite
films of my life . it made thriller a few laughs of new}\tabularnewline
{\scriptsize{}33.33\%} & {\scriptsize{}seriously imdb many previous reviews posted here about
this movie due to a pair of art killings \& <UNK> combined }\tabularnewline
{\scriptsize{}33.33\%} & {\scriptsize{}i haven't been such a clever flick , which is neither
so <UNK> a mystery . the acting is somewhere }\tabularnewline
{\scriptsize{}33.33\%} & {\scriptsize{}so is a lovely film by any <UNK> . it makes a dumb decision
to play fairly typical thing . }\tabularnewline
{\scriptsize{}38.33\%} & {\scriptsize{}wow surprised how wanted to see this . i thought about
two of steven johnson are superb . however , }\tabularnewline
{\scriptsize{}44.44\%} & {\scriptsize{}the tense 1931 melodrama is one of the finest entries
to all ideas . while this is a great anime }\tabularnewline
{\scriptsize{}44.44\%} & {\scriptsize{}i believe i know things i can say , you always come
out of my time and the way max }\tabularnewline
{\scriptsize{}44.44\%} & {\scriptsize{}the movie is a piece of crap . a group of guy lives
off the shelf who meets shrek , }\tabularnewline
{\scriptsize{}50.00\%} & {\scriptsize{}this izzard is a one-in-a-million comic genius ! many
stars in this film ? <br / ><br / <UNK> the }\tabularnewline
\end{tabular}
\end{table}

\subsection*{Variational Autoencoder (VAE)}

Variational Autoencoders (VAEs) \cite{kingma2014auto} approximate
the maximum log-likelihood and can be trained using gradient decent.
VAEs are trained to maximize a variational lower bound $L$ on log-likelihood:

\begin{align*}
L(x;\,\theta) & =\E{}_{z\sim q(z\vert x)}[\log p_{model}(x\vert z)]-D_{KL}[q(z\vert x)\Vert p_{model}(z)]
\end{align*}

where $q(z\vert x)$ is a posterior and $p_{model}(z)$ is a prior
distributions for latent variable $z$. The first term is the data
reconstruction likelihood. The second term works a regularizer to
make $q(z\vert x)$ and $p_{model}(z)$ close to each other. $p_{model}(z)$
can be chosen as $\mathcal{N}(0,\,I)$. $p_{model}(x\vert z)$ is
the decoder, modeled as a neural network that resembles reconstruction
of $x$ from $z$ sampled from the learned $q(z\vert x)$.

\subsection*{Proof of final objective function}

Consider this optimization problem:
\begin{align}
\max_{G}\min_{D}\, & \left[\mathbb{E}_{X\sim p_{d}}\left[\log\,p_{G}\left(X\mid\theta\right)\right]-\text{\ensuremath{\mathbb{E}}}_{X\sim p_{d}}\left[\log\,D\left(X\right)\right]-\mathbb{E}_{z\sim p_{z}}\left[\log\left[1-D\left(G\left(z\right)\right)\right]\right]\right]\label{eq:op_ARN-1}
\end{align}
Given a generator $G$, the optimal $D^{*}\left(G\right)$ is determined
as:
\[
D_{G}^{*}\left(X\right)=\frac{p_{d}\left(X\right)}{p_{G}\left(X\right)+p_{d}\left(X\right)}
\]
where $p_{G}\left(X\right)$ is the distribution induced from $G\left(X\right)$
where $X\sim p_{d}\left(X\right)$.

Substituting $D_{G}^{*}$ back to Eq. (\ref{eq:op_ARN-1}), we obtain
the following optimization problem regarding $G$:
\begin{equation}
\max_{G}\left(\mathbb{E}_{p_{d}}\left[\log\,p_{G}\left(X\right)\right]-I_{JS}\left(P_{d}\Vert P_{G}\right)\right)\label{eq:ARN_G}
\end{equation}

The objective function in Eq. (\ref{eq:ARN_G}) can be written as
\begin{align*}
\mathbb{E}_{p_{d}}\left[\log\,p_{G}\left(X\right)\right]-I_{JS}\left(P_{d}\Vert P_{G}\right) & =-I_{JS}\left(P_{d}\Vert P_{G}\right)-I_{KL}\left(P_{d}\Vert P_{G}\right)-\mathbb{E}_{p_{d}}\left[\log\,p_{d}\right]\\
= & -I_{JS}\left(P_{d}\Vert P_{G}\right)-I_{KL}\left(P_{d}\Vert P_{G}\right)+\text{const}
\end{align*}

Therefore, the optimization problem in Eq. (\ref{eq:ARN_G}) is equivalent
to:
\[
\min_{G}\left(I_{JS}\left(P_{d}\Vert P_{G}\right)+I_{KL}\left(P_{d}\Vert P_{G}\right)\right)
\]

At the Nash equilibrium point of this game, we hence obtain: $p_{G}\left(X\right)=p_{d}\left(X\right)$.

\end{document}